\documentclass[journal]{IEEEtran}
\usepackage{graphicx}
\usepackage{amsmath}   
\usepackage{amssymb}   
\usepackage{amsfonts}  
\usepackage{xcolor}
\usepackage{booktabs}  
\usepackage{multirow}  
\usepackage{makecell}
\usepackage{cite}
\usepackage[hidelinks]{hyperref}
\usepackage{amsmath}
\usepackage{algorithm}
\usepackage{algorithmic}
\usepackage{spreadtab}
\usepackage{numprint}
\usepackage[capitalize]{cleveref}
\usepackage{xcolor}
\title{Residual Gaussian Splatting for Ultra Sparse-View CBCT Reconstruction}

\author{Jian Lin, Jiancheng Fang, Shaoyu Wang, Changan Lai, Yikun Zhang, Yang Chen, \IEEEmembership{Senior Member, IEEE}, and Qiegen Liu, \IEEEmembership{Senior Member, IEEE}
    \thanks{This work was supported by the National Natural Science Foundation of China (U24A20304), and in part by the National Key Research and Development Program of China (Grant 2023YFF1204300 and Grant 2023YFF1204302). (Jian Lin and Jiancheng Fang are co-first authors.) (Corresponding author: Qiegen Liu.)}
    \thanks{J. Lin, J. Fang, C. Lai, S. Wang, and Q. Liu are with the School of Information Engineering, Nanchang University, Nanchang 330031, China (e-mail: liuqiegen@ncu.edu.cn).}
    \thanks{Y. Zhang and Y. Chen are with the Laboratory of Image Science and Technology, the School of Computer Science and Engineering, and the Key Laboratory of New Generation Artificial Intelligence Technology and Its Interdisciplinary Applications, Ministry of Education, Southeast University, Nanjing 210096, China (e-mail: yikun@seu.edu.cn; chenyang.list@seu.edu.cn).}
}
\date{January 2026}

\begin{document}

\maketitle

\begin{abstract}
While 3D Gaussian splatting (3DGS) offers explicit and efficient scene representations for cone-beam computed tomography reconstruction, conventional photometric optimization inherently suffers from spectral bias under ultra sparse-view conditions, leading to over-smoothing and a loss of high-frequency anatomical details. Since wavelet transforms provide rich high-frequency information and have been widely utilized to enhance sparse reconstruction, this work integrates wavelet multi-resolution analysis with 3DGS. To circumvent the mathematical mismatch between the strict non-negativity of physical X-ray attenuation and the bipolar nature of high-frequency wavelet coefficients, we propose Residual Gaussian Splatting (RGS). Methodologically, we introduce a spectrally-decoupled Gaussian representation that stratifies the volumetric field into a geometric base component and a residual detail component. This decomposition systematically transforms explicit high-frequency fitting into a physically consistent, implicit residual compensation task. Furthermore, we devise a spectral-spatial collaborative optimization strategy to coordinate the interplay between geometric anchoring and texture refinement, effectively preventing spectral crosstalk. Extensive experiments on clinical datasets demonstrate that RGS enables the reconstructed images to capture highly refined geometric textures. It successfully resolves the trade-off between artifact suppression and detail preservation, yielding superior visual fidelity in complex trabecular and vascular structures compared to existing neural rendering baselines.
\end{abstract}


\section{Introduction} 
\begin{figure}[htbp]
    \centering
    \includegraphics[width=\columnwidth]{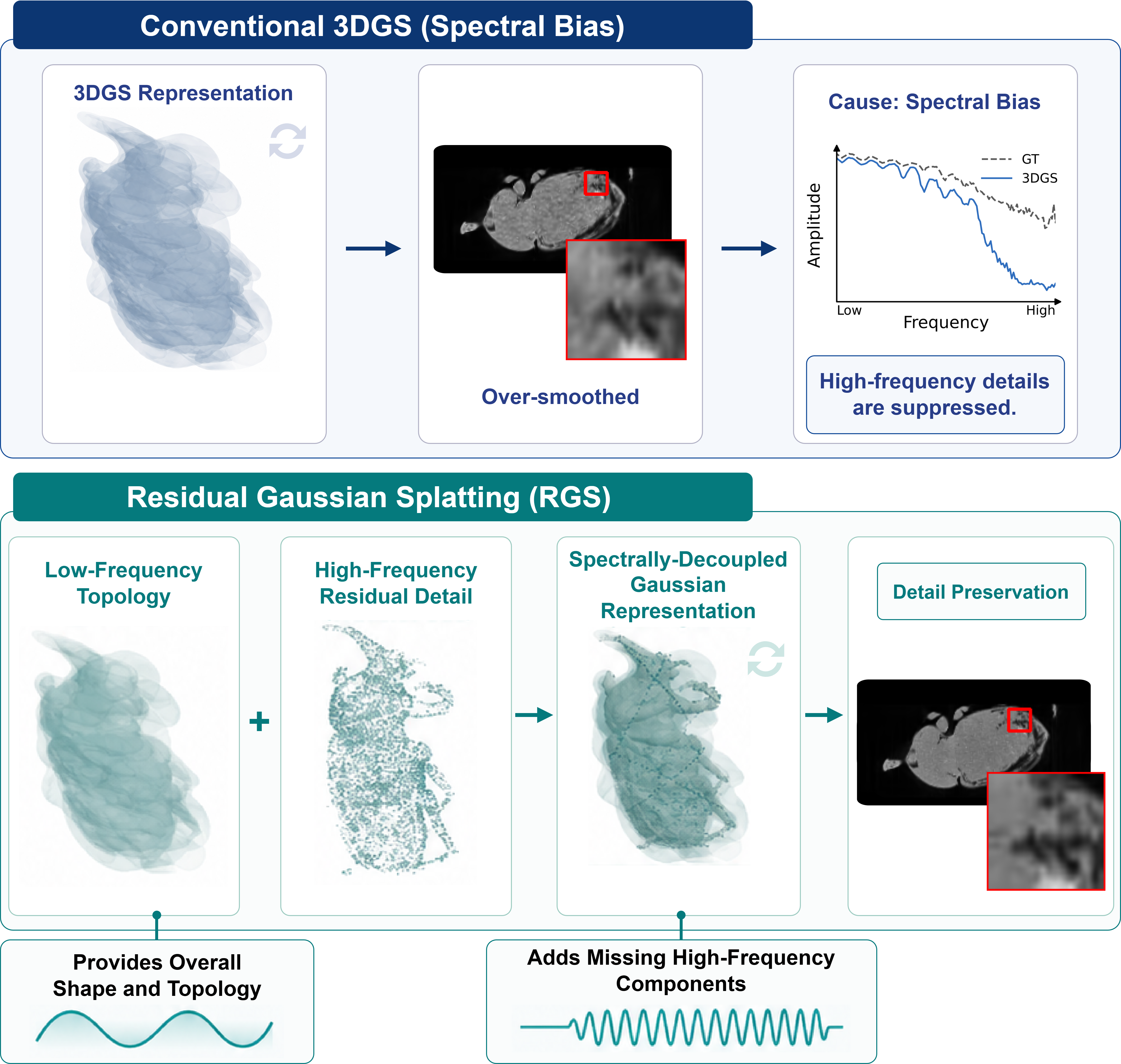}
    \vspace{-15pt}
    \caption{Schematic illustration of the motivation and core concept of the proposed RGS framework, contrasting the over-smoothed reconstructions of standard 3DGS with our spectrally-decoupled Gaussian representation.}
    \label{fig:intro_overview}
    \vspace{-17pt}
\end{figure}

Cone-beam computed tomography (CBCT) is a vital 3D imaging modality in clinical diagnostics. It acquires volumetric data by synchronously rotating a cone-shaped X-ray source and a flat-panel detector around the subject, generating 2D planar projections that are subsequently reconstructed into 3D volumes via mathematical algorithms.

Historically, analytical methods provide computational efficiency but suffer from severe artifacts in sparse-view scenarios. Iterative and Compressed Sensing (CS) approaches mitigate this via sparsity priors \cite{1614066, lustig2007sparse, xuLowDoseXrayCT2012, aharonKSVDAlgorithmDesigning2006}, utilizing tools like multi-scale wavelets for edge preservation \cite{mallatTheoryMultiresolutionSignal1989}. Nevertheless, these methods remain computationally demanding and often produce over-smoothed textures due to excessive regularization.

Recently, deep learning has advanced CT reconstruction through various paradigms \cite{wangDeepLearningTomographic2020a, chenDeepLearningbasedAlgorithms2024}. Projection-domain methods directly restore sinogram consistency \cite{10.1007/978-3-319-46726-9_50, zhangConvolutionalNeuralNetwork2018, leeDeepNeuralNetworkBasedSinogramSynthesis2019}, demonstrating effectiveness in challenging scanning conditions \cite{xuStagebyStageWaveletOptimization2024, guanGenerativeModelingSinogram2024,liDualDomainCollaborativeDiffusion2024}. Conversely, image-domain post-processing reduces artifacts \cite{huHybridDomainNeuralNetwork2021} but risks irreversible information loss and relies heavily on paired data \cite{wolterinkGenerativeAdversarialNetworks2017}. Dual-domain architectures address these shortcomings by enforcing cross-space consistency \cite{zhouDuDoRNetLearningDualDomain2020, linDuDoNetDualDomain2019}, albeit incurring prohibitive computational overhead \cite{wangDeepLearningTomographic2020a, chunMomentumNetFastConvergent2023}. Furthermore, emerging generative paradigms mitigate ill-posedness by implicitly learning complex data distributions for high-fidelity synthesis.

Concurrently, implicit neural representations (INRs) like NeRF \cite{10.1145/3503250} and 3DGS \cite{kerbl3DGaussianSplatting2023} have advanced continuous volumetric modeling. Adapting these to CT, methods such as NAF \cite{10.1007/978-3-031-16446-0_42} and X-GS \cite{10.1007/978-3-031-73232-4_16} utilize implicit fields and 3D Gaussians for differentiable projection. To further improve 3DGS fidelity, recent studies have refined rendering physics to adhere to the Beer-Lambert law \cite{zhaR$^2$GaussianRectifyingRadiative2024} and introduced coarse-to-fine training pipelines \cite{li3DGRCTSparseviewCT2025}.

Despite these advancements, strictly gradient-based optimization inherently suffers from spectral bias \cite{rahamanSpectralBiasNeural2019}, disproportionately favoring low-frequency convergence. In CT reconstruction, global photometric losses are dominated by high-energy structural approximations. Consequently, the optimization overlooks high-frequency, fine-grained anatomical details, yielding the over-smoothed reconstructions characteristic of standard 3DGS, as illustrated in \cref{fig:intro_overview}.

Direct application of wavelet analysis for detail recovery is mathematically hindered because the zero-mean, bipolar nature of high-frequency coefficients contradicts the strictly non-negative X-ray attenuation. To resolve this, we propose Residual Gaussian Splatting (RGS). This framework introduces a spectrally-decoupled Gaussian representation that stratifies the volumetric field into a geometric base component and a residual detail component. As conceptualized in \cref{fig:intro_overview}, superimposing the high-frequency residual onto the low-frequency topology explicitly preserves structural details. Furthermore, a spectral-spatial collaborative optimization strategy reformulates the direct fitting of signed signals into an implicit, non-negative residual compensation task, successfully reconciling global structural consistency with local textural fidelity.

The main contributions of this work are summarized as follows:

\begin{itemize}
    \item \textit{Residual Gaussian Splatting Framework:} We propose a novel reconstruction framework featuring a spectrally-decoupled Gaussian representation. By stratifying the volumetric field into a geometric base and a residual detail component, it fundamentally resolves the mathematical incompatibility between strictly non-negative physical X-ray attenuation and the zero-mean, signed characteristics of high-frequency wavelet coefficients.
    
    \item \textit{Implicit Spectral Compensation Mechanism:} We reformulate the recovery of fine anatomical details as an implicit, non-negative residual compensation task rather than explicit high-frequency fitting. This paradigm strictly adheres to X-ray transmission physics, preventing the optimization instability typically induced by direct spectral supervision.

    \item \textit{Spectral-Spatial Collaborative Optimization:} We devise a curriculum-based strategy orchestrating the interplay between geometric anchoring and texture refinement. By enforcing a low-frequency warm-up phase followed by a frequency-locked constraint, it mitigates spectral crosstalk and ensures robust convergence for complex topological structures without introducing additional inference overhead.
\end{itemize}

The remainder of this paper is organized as follows. Section II reviews the theoretical foundations of CBCT imaging, 3DGS, and wavelet analysis. Section III details the proposed RGS framework. Section IV presents the experimental evaluations, ablation studies, and discussions, followed by the conclusion in Section V.

\section{Related Work}

\subsection{CBCT Imaging System and Geometry}

\begin{figure}[htbp]
    \centering
    \includegraphics[width=\columnwidth]{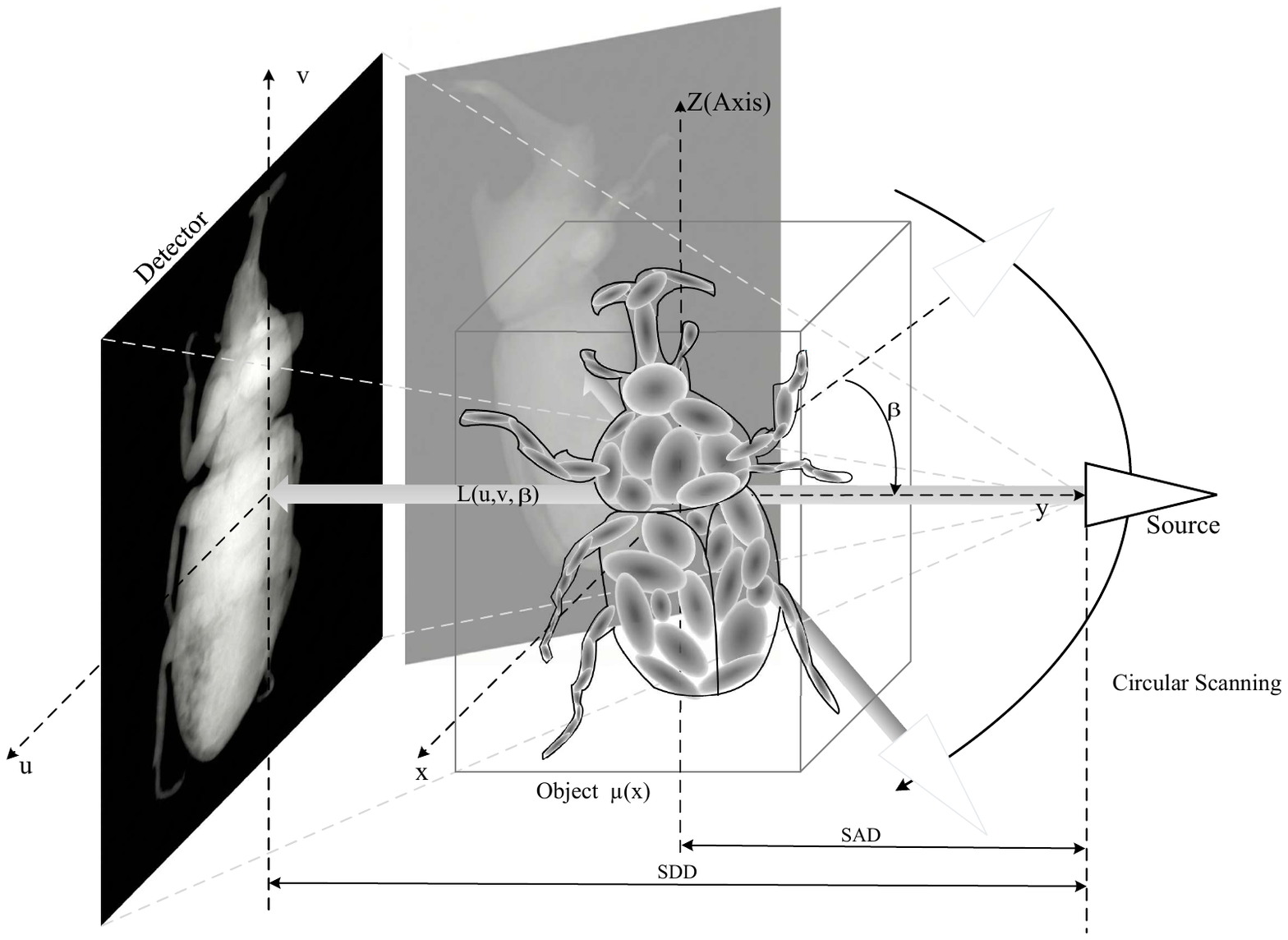} 
    \vspace{-17pt}
    \caption{Schematic illustration of the CBCT imaging geometry.}
    \label{fig:cbct_geometry}
    \vspace{-5pt}
\end{figure}

CBCT employs a cone-shaped X-ray source and a flat-panel detector for volumetric imaging \cite{scarfeWhatConeBeamCT2008, scarfe2006clinical}. As illustrated in \cref{fig:cbct_geometry}, a single synchronized gantry rotation acquires 2D projections that record X-ray attenuation line integrals for 3D reconstruction.

The acquisition geometry is modeled as a pinhole camera defined by the source-to-axis (SAD) and source-to-detector (SDD) distances. Let $\mathbf{x} = (x, y, z)^T \in \mathbb{R}^3$ denote the object coordinates aligned with the rotation axis $z$. For a gantry angle $\beta$, the projection intensity $P(u, v, \beta)$ follows the Beer-Lambert law as the line integral of the linear attenuation coefficient $\mu(\mathbf{x})$:

\begin{equation}
    P(u, v, \beta) = \ln \left( \frac{I_0}{I(u, v, \beta)} \right) = \int_{L(u, v, \beta)} \mu(\mathbf{x}) \, d\mathbf{x}, 
\end{equation}
where $L(u, v, \beta)$ denotes the source-to-pixel ray trajectory \cite{feldkampPracticalConebeamAlgorithm1984}.

\vspace{-10pt}
\subsection{3DGS in CT Reconstruction}

\subsubsection{Fundamentals of 3DGS}

3DGS explicitly models volumetric geometry using a sparse set of anisotropic Gaussian primitives \cite{kerbl3DGaussianSplatting2023}. Each primitive encapsulates an opacity scalar $\alpha$, view-dependent appearance coefficients, and spatial parameters including a center $\mathbf{c} \in \mathbb{R}^3$ and a 3D covariance matrix $\Sigma$, which define its spatial distribution $G(\mathbf{x})$:

\begin{equation}
    G(\mathbf{x}) = \exp \left( -\frac{1}{2} (\mathbf{x} - \mathbf{c})^T \Sigma^{-1} (\mathbf{x} - \mathbf{c}) \right).
\end{equation}

For rendering, EWA splatting \cite{zwickerEWAVolumeSplatting2001} projects these 3D Gaussians to derive a 2D covariance $\Sigma' = J \Sigma J^T$, where $J$ is the Jacobian of the affine view transformation. A tile-based differentiable rasterizer then accumulates the projected contributions via $\alpha$-blending, facilitating efficient end-to-end parameter optimization via gradient descent.

\subsubsection{Adaptation to CT Reconstruction}
\label{CT_reconstruciton_adaptation}

Adapting 3DGS for CT reconstruction shifts the focus from opaque radiance modeling to reconstructing a translucent attenuation field $\mu(\mathbf{x})$. Gaussian primitives are thus repurposed as continuous basis functions to approximate physical density:

\vspace{-8pt}
\begin{equation}
    \mu(\mathbf{x}) = \sum_{i=1}^{N} \sigma_i G(\mathbf{x} | \mathbf{c}_i, \Sigma_i),
\end{equation}
where $\sigma_i$ and $\mathbf{c}_i$ are the density contribution and center of the $i$-th Gaussian. Since tissue X-ray attenuation is isotropic, view-dependent spherical harmonics are discarded \cite{10.1007/978-3-031-73232-4_16}.

Furthermore, rendering must adhere to X-ray transmission physics. Replacing occlusion-based $\alpha$-blending, the projection model strictly follows the additive Beer-Lambert law. For a ray $\mathbf{r}$ with trajectory $L(\mathbf{r})$, the projection intensity $P(\mathbf{r})$ is the line integral of $\mu(\mathbf{x})$:
\vspace{-5pt}
\begin{equation}
    \begin{split}
        P(\mathbf{r}) &\equiv P(u, v, \beta) = \int_{L(\mathbf{r})} \mu(\mathbf{x}) d\mathbf{x} \\
        &\approx \sum_{i \in \mathcal{N}(\mathbf{r})} \sigma_i \int_{L(\mathbf{r})} G(\mathbf{x} | \mathbf{c}_i, \Sigma_i) d\mathbf{x},
    \end{split}
    \vspace{-12pt}
\end{equation}
where $\mathcal{N}(\mathbf{r})$ denotes the subset of primitives intersecting $L(\mathbf{r})$.

This explicit representation ensures superior geometric consistency and computational efficiency. Because the line integral of a 3D Gaussian analytically yields a 2D Gaussian \cite{10.1007/978-3-031-73232-4_16}, it bypasses the discretization errors and ray-marching overhead of implicit models \cite{10.1007/978-3-031-16446-0_42}, significantly accelerating rendering \cite{kerbl3DGaussianSplatting2023}. Recent advances further address rasterizer ``integration bias'' via rectified blending \cite{zhaR$^2$GaussianRectifyingRadiative2024} or geometric regularization \cite{li3DGRCTSparseviewCT2025}. Although this differentiable pipeline efficiently frames reconstruction as a gradient-based optimization, it inherently suffers from spectral bias \cite{rahamanSpectralBiasNeural2019}. The optimization trajectory is dominated by low-frequency structural approximations, fundamentally restricting the capacity to resolve fine-grained, high-frequency anatomical details.

\vspace{-10pt}
\subsection{Wavelet Transform}

The discrete wavelet transform (DWT) decomposes signals into low-frequency approximations and zero-mean high-frequency details \cite{mallatWaveletTourSignal1999}, isolating global topology from local textures. This multi-resolution sparsity provides a foundational prior in compressed sensing CT \cite{lustig2007sparse}, widely utilized across iterative \cite{doi:10.1137/080716542, stephenDistributedOptimizationStatistical2011, https://doi.org/10.1155/2013/185750, chenPriorImageConstrained2008, niuSparseviewXrayCT2014, sandinoCompressedSensingResearch2020} and deep learning frameworks \cite{huHybridDomainNeuralNetwork2021, zhangISTANetInterpretableOptimizationInspired2018, miletoStateoftheArtDeepLearning2024} to preserve anatomical fidelity. However, integrating wavelet analysis into explicit neural rendering remains underexplored. Although 3DGS offers robust spatial parameterization, its gradient-based optimization inherently suffers from spectral bias \cite{rahamanSpectralBiasNeural2019}, resulting in the over-smoothing of fine-grained structures. To address this limitation, we exploit the theoretical complementarity between 3DGS geometry and wavelet regularization. Embedding multi-resolution constraints directly into the physical rendering pipeline formulates a novel spectral-spatial paradigm for high-fidelity CBCT reconstruction.

\section{Method}

\begin{figure}[htbp]
    \centering
    \includegraphics[width=\columnwidth]{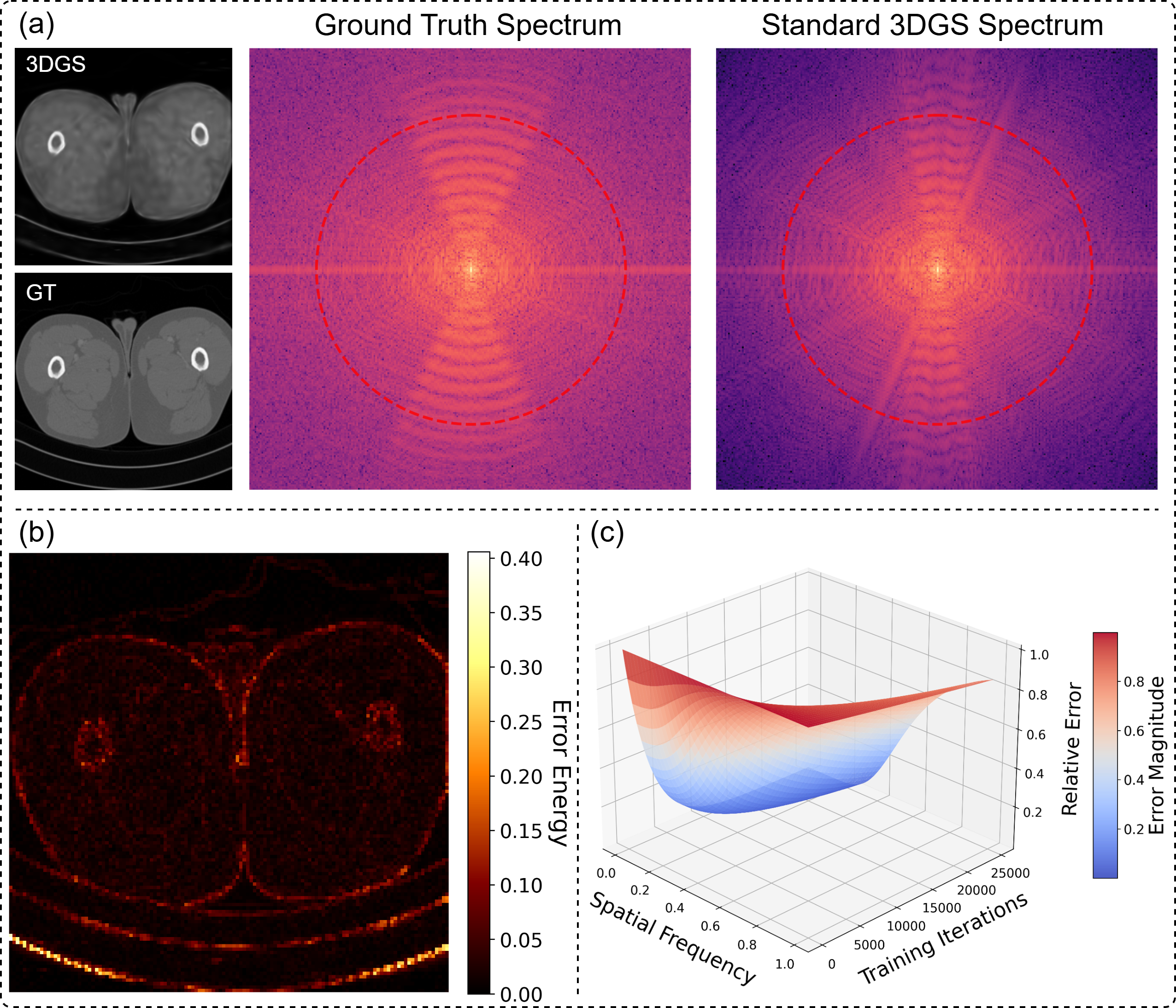}
    \vspace{-17pt}
    \caption{Visual and quantitative analysis demonstrating the inherent spectral bias of standard 3DGS in sparse-view CBCT reconstruction. 
    (a) Spatial reconstructions and their corresponding 2D Fourier magnitude spectra. 
    (b) High-frequency error energy map derived via wavelet decomposition. 
    (c) Optimization dynamics of the relative error across spatial frequencies over training iterations.}
    \label{fig:motivation_figure}
    \vspace{-10pt}
\end{figure}

\begin{figure*}[htbp]
    \centering
    \includegraphics[width=\textwidth]{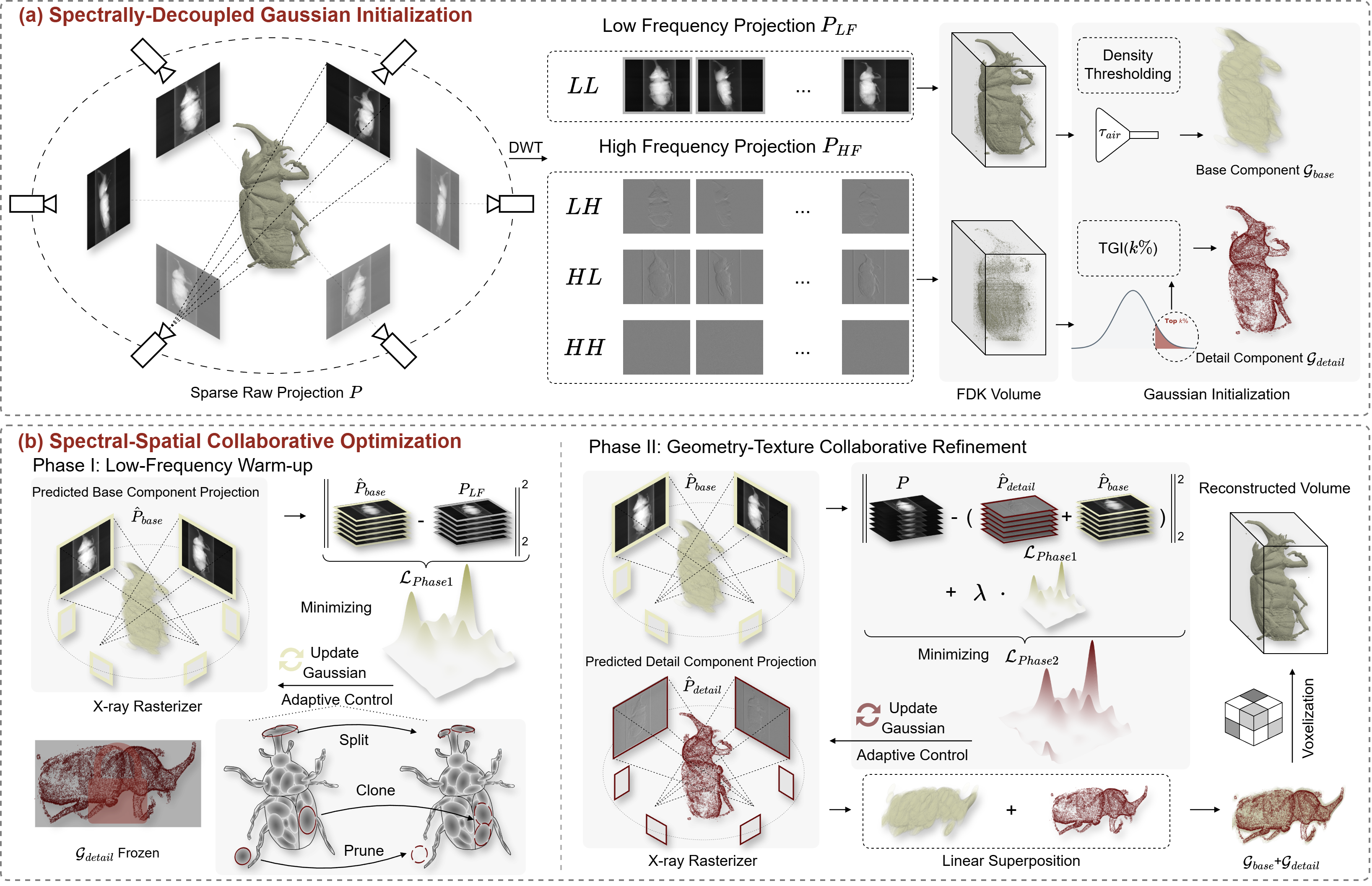}
    \vspace{-18pt}
    \caption{Schematic overview of the proposed RGS framework. 
    (a) Spectrally-decoupled Gaussian initialization. Sparse raw projections are decomposed via DWT to instantiate the base and detail components utilizing low- and high-frequency priors, respectively. 
    (b) Spectral-spatial collaborative optimization. A two-phase training schedule comprising an initial low-frequency warm-up for the base component, followed by a joint geometry-texture refinement phase to implicitly compensate for high-frequency residuals.}
    \label{fig:framework_overview}
    \vspace{-8pt}
\end{figure*}

\subsection{Motivation}
\label{sec:motivation}

While 3D Gaussian representations offer efficient and continuous volumetric modeling for CT reconstruction, standard projection-driven optimization inherently struggles to preserve fine anatomical details due to severe spectral bias. As illustrated in \cref{fig:motivation_figure}(a), 3DGS captures low-frequency global topology but severely depletes high-frequency energy relative to the ground truth. This frequency-domain discrepancy localizes along structural boundaries, as shown in \cref{fig:motivation_figure}(b), and stems directly from the optimization dynamics in \cref{fig:motivation_figure}(c), where low-frequency errors converge rapidly while high-frequency relative errors persistently plateau. This frequency imbalance inevitably yields over-smoothed textures. This fundamental limitation motivates our integration of multi-resolution DWT analysis. By explicitly separating low-frequency structural information from zero-mean high-frequency details, the DWT provides a complementary prior that allows the framework to isolate and preserve local textures without compromising the global geometric anchor modeled by 3DGS.

\vspace{-8pt}

\subsection{Architecture of RGS}
\label{sec:framework_overview}

RGS reconstructs a 3D attenuation field from sparse-view X-ray projections by combining spectral prior decomposition with dual-component Gaussian splatting. As illustrated in \cref{fig:framework_overview}, the pipeline consists of two sequential stages: (a) spectrally-decoupled Gaussian initialization and (b) spectral-spatial collaborative optimization. In the first stage, frequency-specific priors are extracted from the input projections to initialize two Gaussian components. In the second stage, these components are refined directly in the raw projection domain using a two-phase optimization schedule. After optimization, the learned continuous Gaussian field is voxelized on a predefined Cartesian grid to obtain the final reconstructed volume.

Let $P=\{P^{v}\}_{v=1}^{V}$ denote the sparse-view projection set, where $V$ is the number of acquisition views and $P^{v}$ is the projection at the $v$-th view, we first apply a single-level 2D DWT to each projection:
\begin{equation}
\begin{aligned}
    (P_{LF}, P_{LH}, P_{HL}, P_{HH}) = \mathrm{DWT}(P),
\end{aligned}
\label{eq:framework_init_1}
\end{equation}
which produces one low-frequency approximation sub-band and three directional high-frequency sub-bands. The low-frequency component is analytically reconstructed to provide a coarse geometric prior, $V_{LF} = \mathrm{FDK}(P_{LF})$, where $\mathrm{FDK}(\cdot)$ denotes the Feldkamp--Davis--Kress reconstruction operator.

In parallel, the three high-frequency sub-bands are aggregated into a non-negative high-frequency energy map:
\begin{equation}
\begin{aligned}
    M_{HF} = |P_{LH}| + |P_{HL}| + |P_{HH}|,
\end{aligned}
\label{eq:framework_init_3}
\end{equation}
which is then backprojected as $V_{sal} = \mathcal{B}(M_{HF})$, where $M_{HF}$ is the high-frequency energy map, $V_{sal}$ is the volumetric high-frequency prior, and $\mathcal{B}(\cdot)$ denotes the backprojection operator. Importantly, the high-frequency coefficients are not used as direct supervision targets during rendering. Instead, they are only used to construct a non-negative spatial prior for detail initialization, while the actual recovery of fine structures is completed later through residual fitting in the raw projection domain. The rationale for this design is discussed in Section~\ref{sec:spectral_init}.

Based on these two priors, we initialize two Gaussian components:
\begin{equation}
    \mathcal{G}_{base} = \Phi_{base}(V_{LF};\tau_{air}), \quad \mathcal{G}_{detail} = \Phi_{detail}(V_{sal};k),
    \label{eq:framework_init_4}
\end{equation}
where $\mathcal{G}_{base}$ and $\mathcal{G}_{detail}$ denote the base and detail Gaussian sets, respectively. $\Phi_{base}(\cdot)$ denotes density-thresholded initialization from the low-frequency reconstruction, with $\tau_{air}$ used to suppress the air region. $\Phi_{detail}(\cdot)$ denotes saliency-guided initialization from the high-frequency prior, and $k$ controls the proportion of high-saliency locations used for detail initialization.

After initialization, the attenuation field is represented as the superposition of the two Gaussian components:
\begin{equation}
\begin{aligned}
    \hat{\mu}(\mathbf{x})
    =
    \sum_{i \in \mathcal{I}_{base}} \sigma_i G(\mathbf{x}\mid \mathbf{c}_i,\Sigma_i)
    +
    \sum_{j \in \mathcal{I}_{detail}} \sigma_j G(\mathbf{x}\mid \mathbf{c}_j,\Sigma_j),
\end{aligned}
\label{eq:framework_representation_1}
\end{equation}
where $\mathbf{x}\in\mathbb{R}^{3}$ denotes the spatial coordinate, $\mathcal{I}_{base}$ and $\mathcal{I}_{detail}$ are the index sets of the two components, and $\theta_n=\{\mathbf{c}_n,\Sigma_n,\sigma_n\}$ denotes the parameters of the $n$-th Gaussian primitive, including its center $\mathbf{c}_n$, covariance matrix $\Sigma_n$, and density coefficient $\sigma_n$. Here, $G(\mathbf{x}\mid \mathbf{c}_n,\Sigma_n)$ denotes the Gaussian basis function.

The corresponding forward projections are obtained by differentiable rendering, where $\mathcal{R}(\cdot)$ maps $\mathcal{G}_{base}$ and $\mathcal{G}_{detail}$ to the component-wise projections $\hat{P}_{base}$ and $\hat{P}_{detail}$, whose sum forms the synthesized projection $\hat{P}$.

The optimization is performed in two phases. In Phase~I, only the base component is optimized against the low-frequency target to stabilize the global anatomical topology, while the detail component remains frozen. In Phase~II, both components are jointly refined in the raw projection domain under a combined objective $\mathcal{L}=\mathcal{L}_{Phase1}+\mathcal{L}_{global}+\lambda\mathcal{L}_{cons}$, where $\mathcal{L}_{Phase1}$ denotes the low-frequency fitting loss between $P_{LF}$ and the rendered base projection, $\mathcal{L}_{global}$ denotes the projection fidelity loss between $P$ and the synthesized projection, and $\mathcal{L}_{cons}$ denotes the low-frequency consistency loss between $P_{LF}$ and the rendered base projection during joint refinement. The detailed phase-wise formulations are given in Section~\ref{sec:collaborative_optimization}. In both phases, gradient-based updates are coupled with adaptive density control to split, clone, or prune Gaussian primitives when needed.

After optimization, the learned Gaussian representation defines a continuous attenuation field, which is voxelized on a predefined Cartesian grid $\Omega=\{\mathbf{x}_m\}_{m=1}^{M}$ by evaluating $\hat{V}[m]=\hat{\mu}(\mathbf{x}_m)$ for $\mathbf{x}_m\in\Omega$, where $\hat{V}$ denotes the reconstructed 3D attenuation volume. Therefore, RGS first constructs frequency-aware Gaussian priors, then refines them through residual projection fitting, and finally converts the optimized continuous Gaussian field into the volumetric reconstruction used for visualization and quantitative evaluation. The complete execution flow is summarized in Algorithm~\ref{alg:sr_gs}.

\begin{algorithm}[htbp]
    \caption{RGS}
    \label{alg:sr_gs}
    \textbf{Input:} Sparse-view projections $P=\{P^{v}\}_{v=1}^{V}$, voxel grid $\Omega=\{\mathbf{x}_m\}_{m=1}^{M}$, total iterations $T$, Phase-I iterations $T_{warm}$, density threshold $\tau_{air}$, saliency ratio $k$, loss weight $\lambda$, learning rates $\eta_1,\eta_2$ \\
    \textbf{Output:} Reconstructed volume $\hat{V}$
    
    \begin{algorithmic}[1]
        \STATE $(P_{LF}, P_{LH}, P_{HL}, P_{HH}) \leftarrow \mathrm{DWT}(P)$
        \STATE $V_{LF} \leftarrow \mathrm{FDK}(P_{LF})$
        \STATE $M_{HF} \leftarrow |P_{LH}| + |P_{HL}| + |P_{HH}|$
        \STATE $V_{sal} \leftarrow \mathcal{B}(M_{HF})$
        \STATE $(\mathcal{G}_{base}, \mathcal{G}_{detail}) \leftarrow (\Phi_{base}(V_{LF}; \tau_{air}), \Phi_{detail}(V_{sal}; k))$
        
        \vspace{-0.3em}
        \item[] \hspace{-1.1em} \rule{\linewidth}{0.5pt}
        \item[] \hspace{-1.1em} \textbf{Phase I:} Freeze $\mathcal{G}_{detail}$ and optimize $\mathcal{G}_{base}$
        \vspace{-0.5em}
        \item[] \hspace{-1.1em} \rule{\linewidth}{0.5pt}
        
        \FOR{$t = 0$ to $T_{warm}-1$}
            \STATE $\hat{P}_{base} \leftarrow \mathcal{R}(\mathcal{G}_{base})$
            \STATE $\mathcal{L}_{Phase1} \leftarrow \|P_{LF} - \hat{P}_{base}\|_2^2$
            \STATE $\mathcal{G}_{base} \leftarrow \mathrm{ADC}\!\left(\mathcal{G}_{base} - \eta_1 \nabla_{\mathcal{G}_{base}} \mathcal{L}_{Phase1}\right)$
        \ENDFOR
        
        \vspace{-0.3em}
        \item[] \hspace{-1.1em} \rule{\linewidth}{0.5pt}
        \item[] \hspace{-1.1em} \textbf{Phase II:} Jointly refine $\mathcal{G}_{base}$ and $\mathcal{G}_{detail}$
        \vspace{-0.5em}
        \item[] \hspace{-1.1em} \rule{\linewidth}{0.5pt}
        
        \FOR{$t = T_{warm}$ to $T-1$}
            \STATE $(\hat{P}_{base}, \hat{P}_{detail}) \leftarrow (\mathcal{R}(\mathcal{G}_{base}), \mathcal{R}(\mathcal{G}_{detail}))$
            \STATE $\hat{P} \leftarrow \hat{P}_{base} + \hat{P}_{detail}$
            \STATE $(\mathcal{L}_{global}, \mathcal{L}_{cons}) \leftarrow (\|P - \hat{P}\|_2^2, \ \|P_{LF} - \hat{P}_{base}\|_2^2)$
            \STATE $\mathcal{L}_{Phase2} \leftarrow \mathcal{L}_{global} + \lambda \mathcal{L}_{cons}$
            \STATE $\mathcal{G}_{base} \leftarrow \mathrm{ADC}\!\left(\mathcal{G}_{base} - \eta_2 \nabla_{\mathcal{G}_{base}} \mathcal{L}_{Phase2}\right)$
            \STATE $\mathcal{G}_{detail} \leftarrow \mathrm{ADC}\!\left(\mathcal{G}_{detail} - \eta_2 \nabla_{\mathcal{G}_{detail}} \mathcal{L}_{Phase2}\right)$
        \ENDFOR
        
        \vspace{0.4em}
        \STATE \textbf{Voxelization:} Query the optimized Gaussian field on $\Omega$
        \FOR{$m = 1$ to $M$}
            \STATE $\hat{\mu}(\mathbf{x}_m) \leftarrow \displaystyle\sum_{i\in\mathcal{I}_{base}} \sigma_i G(\mathbf{x}_m \mid \mathbf{c}_i, \Sigma_i)$ \\
            \hspace{1cm} $+ \displaystyle\sum_{j\in\mathcal{I}_{detail}} \sigma_j G(\mathbf{x}_m \mid \mathbf{c}_j, \Sigma_j)$
            \STATE $\hat{V}[m] \leftarrow \hat{\mu}(\mathbf{x}_m)$
        \ENDFOR
                    
        \STATE \textbf{return} $\hat{V}$
    \end{algorithmic}
\end{algorithm}

\subsection{Spectrally-Decoupled Gaussian Initialization}
\label{sec:spectral_init}

This subsection explains why spectral decomposition is used only for initialization and how it leads to the dual-component Gaussian representation. In RGS, the low-frequency and high-frequency contents of the attenuation field are assigned different spatial priors before iterative optimization. This design alleviates the spectral bias of direct projection-driven fitting and provides a more stable starting point for the subsequent collaborative refinement.

Following the non-negative attenuation model established in \cref{CT_reconstruciton_adaptation}, both the continuous field and its rendered projection remain strictly non-negative. Conversely, the high-frequency wavelet sub-bands ($P_{LH}^v$, $P_{HL}^v$, $P_{HH}^v$) derived in \cref{eq:framework_init_1} are signed and exhibit near-zero means \cite{mallatTheoryMultiresolutionSignal1989,mallatWaveletTourSignal1999}. Denoting a specific high-frequency sub-band as $P_h^{v}$ where $h\in\{LH,HL,HH\}$, positive and negative values inherently coexist. A mathematical incompatibility therefore arises when attempting to directly supervise the rendered detail term using these signed coefficients. If the explicit high-frequency objective is defined as
\begin{equation}
\mathcal{L}_{HF}^{exp} = \sum_{h \in \{LH,HL,HH\}} \left\|
P_h - \hat{P}_h \right\|_2^2,
\label{eq:explicit_hf_loss}
\end{equation}
then at any detector location $q$ where $P_h(q)<0$, the rendered value $\hat{P}_h(q)$ still cannot become negative. The derivative with respect to the rendered value is $\frac{\partial \mathcal{L}_{HF}^{exp}}{\partial \hat{P}_h(q)} = 2\bigl(\hat{P}_h(q)-P_h(q)\bigr)$. In regions with negative targets, this derivative remains positive, so optimization suppresses the rendered value instead of using it to fit the signed coefficient. This becomes clearer by expressing the rendered value at detector location $q$ as $\hat{P}_h(q) = \sum_{n=1}^{N}\sigma_n K_n(q)$, where $K_n(q)$ denotes the contribution of the $n$-th Gaussian primitive to detector location $q$:
\vspace{-10pt}
\begin{equation}
K_n(q)
=
\int_{L(q)} G(\mathbf{x}\mid \mathbf{c}_n,\Sigma_n)\,d\mathbf{x}.
\label{eq:init_rendered_response_2}
\end{equation}

Because the Gaussian basis is non-negative, $K_n(q)$ is also non-negative. The gradient with respect to the density coefficient is then
\begin{equation}
\frac{\partial \mathcal{L}_{HF}^{exp}}{\partial \sigma_n}
=
2\sum_{q}
\bigl(\hat{P}_h(q)-P_h(q)\bigr)K_n(q).
\label{eq:hf_grad_sigma}
\end{equation}
For locations where the target coefficient is negative, both $\hat{P}_h(q)-P_h(q)$ and $K_n(q)$ remain non-negative. Consequently, the gradient in Eq.~\eqref{eq:hf_grad_sigma} continuously drives $\sigma_n$ toward smaller values during gradient descent. The primitive is therefore suppressed rather than used to represent the desired high-frequency component.

This analysis reveals a fundamental mismatch between signed wavelet supervision and non-negative attenuation rendering, which we refer to as the incompatibility of spectral supervision. Therefore, in RGS, the high-frequency sub-bands are not used as direct supervision targets during rendering. Instead, they are converted into a non-negative high-frequency energy map for initializing $\mathcal{G}_{detail}$, while the actual recovery of fine structures is deferred to residual fitting in the raw projection domain.

\subsection{Spectral-Spatial Collaborative Optimization}
\label{sec:collaborative_optimization}

With $\mathcal{G}_{base}$ and $\mathcal{G}_{detail}$ initialized as above, reconstruction proceeds directly in the projection domain. The attenuation field follows the dual-component Gaussian representation in Eq.~\eqref{eq:framework_representation_1}. For a ray trajectory $L(\mathbf{r})$, the rendered projections are denoted by $\hat{P}_{base}(\mathbf{r})=\mathcal{R}(\mathcal{G}_{base};\mathbf{r})$ and $\hat{P}_{detail}(\mathbf{r})=\mathcal{R}(\mathcal{G}_{detail};\mathbf{r})$, and their sum forms the total projection:
\begin{equation}
\hat{P}(\mathbf{r})
=
\hat{P}_{base}(\mathbf{r})+\hat{P}_{detail}(\mathbf{r}).
\label{eq:additive_projection_main_2}
\end{equation}

As established in Section~\ref{sec:spectral_init}, directly fitting the signed high-frequency wavelet coefficients is incompatible with the non-negative attenuation model. For this reason, the optimization is formulated in the raw projection domain through the global data-fidelity term
\begin{equation}
\mathcal{L}_{global}
=
\|P-(\hat{P}_{base}+\hat{P}_{detail})\|_2^2.
\label{eq:global_projection_loss}
\end{equation}
Under this formulation, $\mathcal{G}_{detail}$ is not required to reproduce the signed wavelet coefficients explicitly. Instead, it learns the residual projection content that remains unexplained after the base component captures the dominant low-frequency structure. The gradient with respect to the detail branch is
\begin{equation}
\nabla_{\mathcal{G}_{detail}}\mathcal{L}_{global}
=
-2\bigl[P-\hat{P}_{base}-\hat{P}_{detail}\bigr]
\frac{\partial \hat{P}_{detail}}{\partial \mathcal{G}_{detail}}.
\label{eq:detail_gradient_main}
\end{equation}
This shows that the update of the detail branch is driven by the remaining projection error after subtracting the base response.

Although Eq.~\eqref{eq:global_projection_loss} resolves the sign inconsistency, it does not by itself guarantee a stable division of labor between the two branches. If $\mathcal{G}_{base}$ and $\mathcal{G}_{detail}$ are optimized jointly from the beginning, the base branch usually dominates the early iterations because it starts from a stronger volumetric prior and has broader spatial support. As a result, it tends to absorb projection errors that should instead be assigned to the detail branch. To avoid this behavior, we adopt a two-phase optimization schedule.

In the first phase, $\mathcal{G}_{detail}$ is frozen and only $\mathcal{G}_{base}$ is optimized with the low-frequency target $P_{LF}$ over the first $T_{warm}$ iterations. The objective is
\begin{equation}
\mathcal{L}_{Phase1}
=
\|P_{LF}-\hat{P}_{base}\|_2^2.
\label{eq:phase1_loss_main}
\end{equation}
This phase allows the base branch to fit the smooth anatomical structure before the detail branch is activated. The corresponding update is
\begin{equation}
\mathcal{G}_{base}^{\,t+1}
=
\mathrm{ADC}\!\left(
\mathcal{G}_{base}^{\,t}
-
\eta_1 \nabla_{\mathcal{G}_{base}}\mathcal{L}_{Phase1}
\right),
\label{eq:phase1_update_main}
\end{equation}
where $\eta_1$ is the learning rate in the first phase and $\mathrm{ADC}(\cdot)$ denotes adaptive density control.

After the first phase, both branches are jointly refined. To preserve the coarse anatomical structure already captured by the base branch, we combine the full projection error with a low-frequency consistency term, yielding the Phase II objective:
\begin{equation}
\mathcal{L}_{Phase2} = \mathcal{L}_{global} + \lambda \|P_{LF} - \hat{P}_{base}\|_2^2,
\label{eq:phase2_loss_main}
\end{equation}
where $\lambda$ controls the strength of the low-frequency consistency constraint. Under this objective, the base branch remains aligned with the low-frequency target, while the detail branch minimizes the residual projection error. Both components are updated via standard gradient descent coupled with adaptive density control. For the base component, the update rule is defined as $\mathcal{G}_{base}^{\,t+1} = \mathrm{ADC}( \mathcal{G}_{base}^{\,t} - \eta_2 \nabla_{\mathcal{G}_{base}}\mathcal{L}_{Phase2} )$, where $\eta_2$ is the learning rate for the second phase. An identical update formulation is applied to $\mathcal{G}_{detail}$.

Through this collaborative optimization, the base branch mainly preserves the global low-frequency structure, while the detail branch incrementally captures the remaining fine-scale information from the projection residual. The final attenuation field is obtained by summing the two components, namely $\hat{\mu}(\mathbf{x})=\hat{\mu}_{base}(\mathbf{x})+\hat{\mu}_{detail}(\mathbf{x})$, thereby recovering global structure and local details within one physically consistent reconstruction framework.

\begin{figure*}[t]
    \centering
    \setlength{\tabcolsep}{0.5pt}
    \renewcommand{\arraystretch}{1.0}
    
    \begin{tabular}{ccccccc}
        \includegraphics[width=0.13\textwidth]{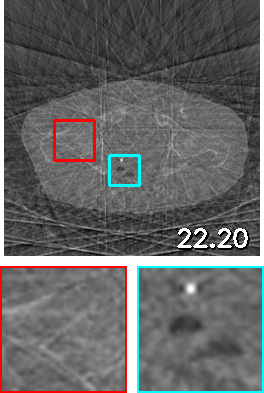} &
        \includegraphics[width=0.13\textwidth]{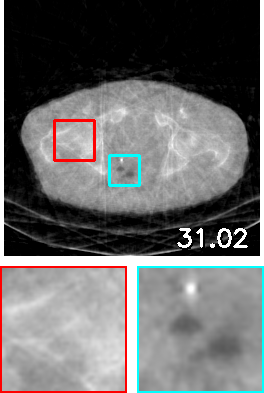} &
        \includegraphics[width=0.13\textwidth]{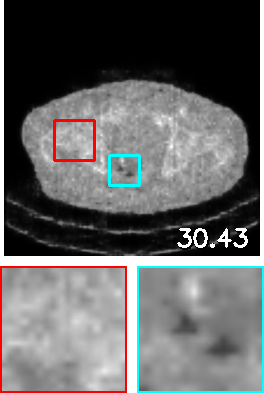} &
        \includegraphics[width=0.13\textwidth]{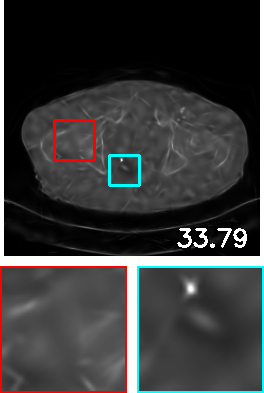} &
        \includegraphics[width=0.13\textwidth]{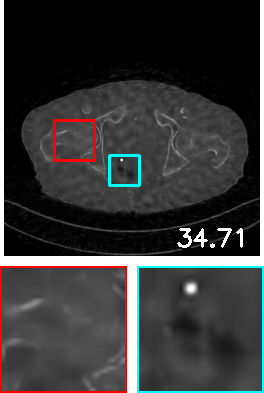} &
        \includegraphics[width=0.13\textwidth]{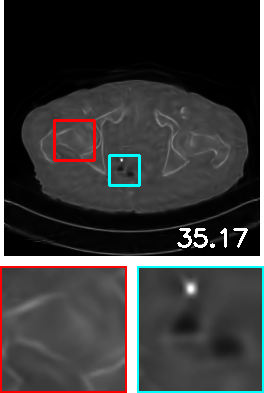} &
        \includegraphics[width=0.13\textwidth]{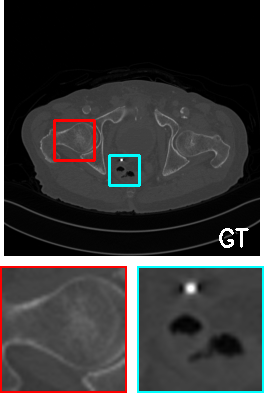} \\
        
        \includegraphics[width=0.13\textwidth]{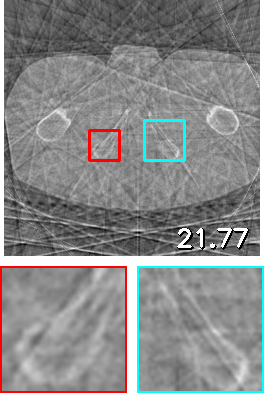} &
        \includegraphics[width=0.13\textwidth]{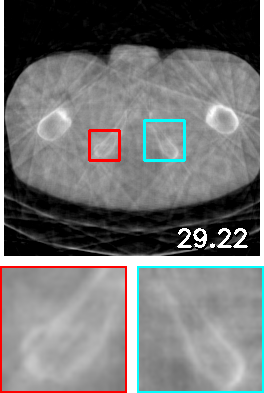} &
        \includegraphics[width=0.13\textwidth]{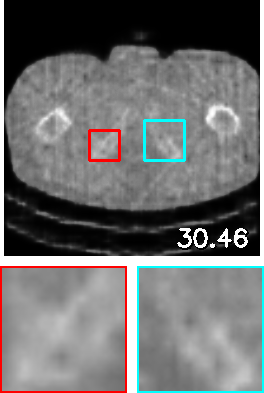} &
        \includegraphics[width=0.13\textwidth]{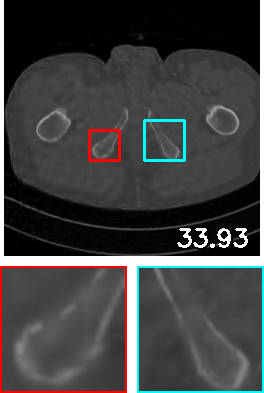} &
        \includegraphics[width=0.13\textwidth]{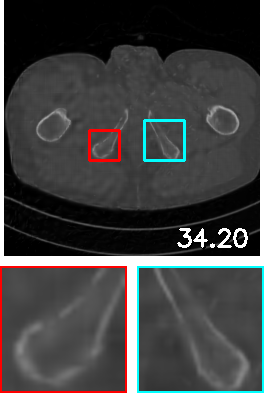} &
        \includegraphics[width=0.13\textwidth]{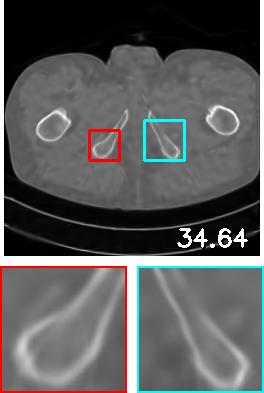} &
        \includegraphics[width=0.13\textwidth]{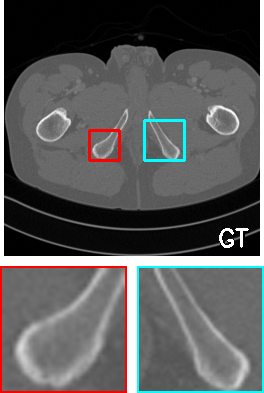} \\
        
        \includegraphics[width=0.13\textwidth]{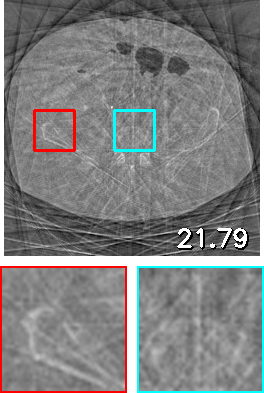} &
        \includegraphics[width=0.13\textwidth]{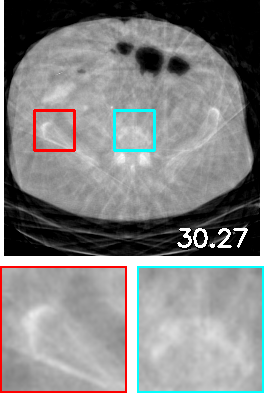} &
        \includegraphics[width=0.13\textwidth]{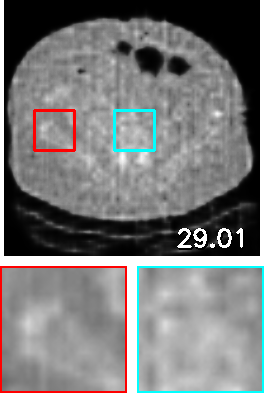} &
        \includegraphics[width=0.13\textwidth]{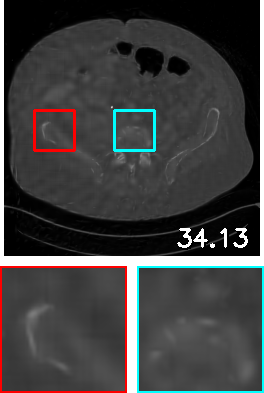} &
        \includegraphics[width=0.13\textwidth]{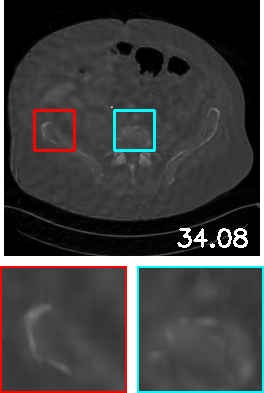} &
        \includegraphics[width=0.13\textwidth]{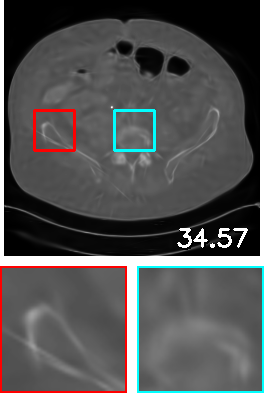} &
        \includegraphics[width=0.13\textwidth]{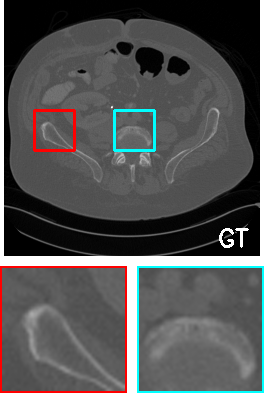} \\

        \includegraphics[width=0.13\textwidth]{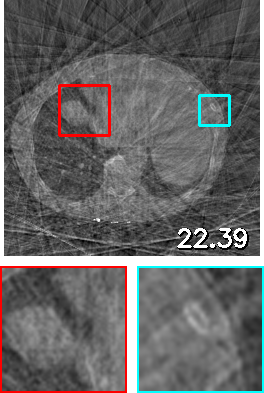} &
        \includegraphics[width=0.13\textwidth]{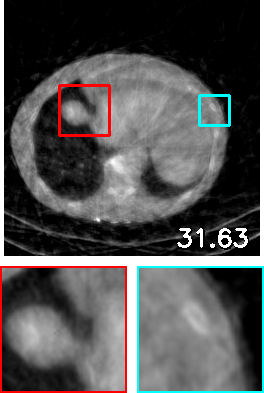} &
        \includegraphics[width=0.13\textwidth]{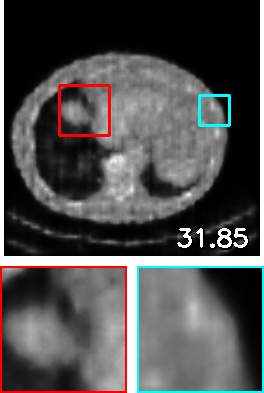} &
        \includegraphics[width=0.13\textwidth]{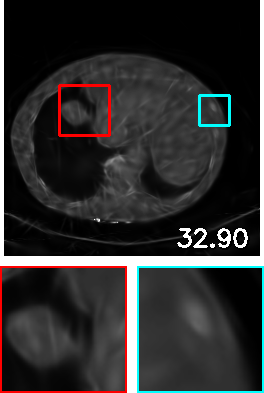} &
        \includegraphics[width=0.13\textwidth]{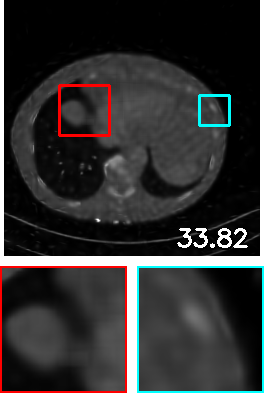} &
        \includegraphics[width=0.13\textwidth]{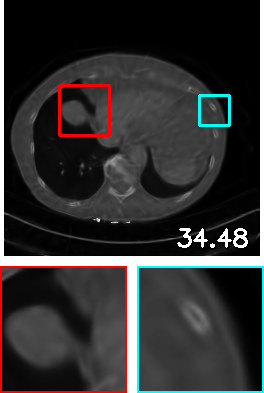} &
        \includegraphics[width=0.13\textwidth]{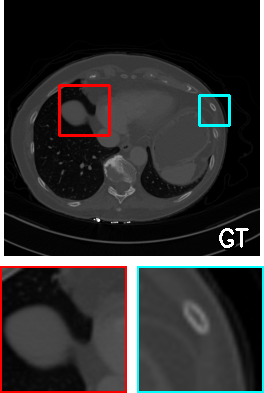} \\
        
        \scriptsize{(a) FDK} &
        \scriptsize{(b) SART} &
        \scriptsize{(c) NAF} &
        \scriptsize{(d) R$^2$-Gaussian} &
        \scriptsize{(e) 3DGR} &
        \scriptsize{(f) Ours} &
        \scriptsize{(g) GT} \\
    \end{tabular}
    
    \setlength{\abovecaptionskip}{10pt}
    
    \vspace{-2mm}
    \caption{Visual comparison of reconstructed chest and abdomen slices from the AAPM dataset under a 20-view sparse configuration. Magnified regions of interest (red and cyan boxes) are provided to illustrate the recovery of fine anatomical details across different reconstruction methods (a)--(f) relative to the GT (g).}
    \label{fig:visual_comparison_20views}
    \vspace{-5mm}
\end{figure*}

\section{Experiments}
\label{sec:experiments}

In this section, we evaluate the RGS framework on the clinical AAPM dataset and real-world biological specimens under varying sparse-view configurations. By benchmarking against representative analytical and state-of-the-art neural rendering methods, we validate both its quantitative reconstruction accuracy and high-frequency detail preservation. Furthermore, ablation studies are provided to verify the respective contributions of the wavelet module to fine detail recovery and the curriculum control strategy to optimization stability. Our source code is publicly available at: https://github.com/yqx7150/RGS.

\subsection{Experimental Setup}
\label{sec:setup}

\begin{table*}[t]
\centering
\caption{Quantitative(PSNR / SSIM) results on the clinical dataset under sparse-view settings. We highlight the \textbf{best} and \underline{second-best} results.}
\label{tab:clinical_results}
\small
\setlength{\tabcolsep}{9pt} 
\renewcommand{\arraystretch}{1.1}
\begin{tabular}{c|c|c|c|c|c|c|c}
\hline\hline
\makecell{\textbf{View}}
& \textbf{Dataset} 
& \textbf{FDK} 
& \textbf{SART} 
& \textbf{NAF} 
& \textbf{R$^2$-Gaussian} 
& \textbf{3DGR} 
& \textbf{Ours} \\
\hline
\multirow{7}{*}{\textbf{20-view}} 
 & L067 & 21.77 / 0.558 & 29.22 / 0.831 & 29.76 / 0.844 & 33.88 / 0.893 & \underline{34.40} / \underline{0.908} & \textbf{34.95} / \textbf{0.914} \\
 & L096 & 21.94 / 0.516 & 30.02 / 0.849 & 30.28 / 0.856 & \underline{33.14} / 0.883 & 32.99 / \underline{0.884} & \textbf{33.53} / \textbf{0.894} \\
 & L109 & 21.57 / 0.499 & 28.89 / 0.822 & 30.01 / 0.850 & 34.02 / \underline{0.902} & \underline{34.49} / 0.901 & \textbf{34.91} / \textbf{0.915} \\
 & L143 & 21.79 / 0.530 & 30.26 / 0.853 & 30.78 / 0.852 & 33.82 / 0.892 & \underline{34.08} / \underline{0.903} & \textbf{34.56} / \textbf{0.911} \\
 & L192 & 22.36 / 0.489 & 31.63 / 0.857 & 32.15 / 0.863 & 34.40 / \underline{0.894} & \underline{34.81} / 0.893 & \textbf{35.26} / \textbf{0.906} \\
 & L506 & 22.20 / 0.577 & 31.09 / 0.871 & 31.83 / 0.864 & 33.71 / 0.886 & \underline{34.13} / \underline{0.891} & \textbf{34.86} / \textbf{0.913} \\
 & Avg. & 21.94 / 0.528 & 30.19 / 0.847 & 30.80 / 0.855 & 33.83 / 0.892 & \underline{34.15} / \underline{0.897} & \textbf{34.68} / \textbf{0.909} \\
\hline\hline
\multirow{7}{*}{\textbf{40-view}} 
 & L067 & 24.58 / 0.625 & 32.29 / 0.933 & 32.64 / 0.924 & 36.64 / 0.945 & \underline{37.17} / \underline{0.957} & \textbf{37.76} / \textbf{0.960} \\
 & L096 & 24.64 / 0.630 & 33.37 / 0.904 & 33.64 / 0.918 & 35.69 / \underline{0.927} & \underline{35.84} / 0.922 & \textbf{36.01} / \textbf{0.932} \\
 & L109 & 24.88 / 0.596 & 31.52 / 0.913 & 33.85 / 0.936 & \underline{37.50} / \underline{0.954} & 37.34 / 0.942 & \textbf{37.88} / \textbf{0.962} \\
 & L143 & 24.69 / 0.649 & 33.97 / 0.919 & 34.21 / 0.927 & 36.07 / 0.930 & \underline{36.73} / \underline{0.933} & \textbf{36.92} / \textbf{0.941} \\
 & L192 & 25.23 / 0.615 & 34.17 / 0.925 & 35.78 / 0.932 & 37.58 / 0.956 & \underline{37.93} / \underline{0.957} & \textbf{38.57} / \textbf{0.960} \\
 & L506 & 25.25 / 0.573 & 34.29 / 0.923 & 35.67 / 0.941 & 37.17 / 0.952 & \underline{38.02} / \underline{0.961} & \textbf{38.41} / \textbf{0.968} \\
 & Avg. & 24.88 / 0.615 & 33.27 / 0.920 & 34.30 / 0.930 & 36.78 / 0.944 & \underline{37.17} / \underline{0.945} & \textbf{37.59} / \textbf{0.954} \\
\hline\hline
\multirow{7}{*}{\textbf{60-view}} 
 & L067 & 25.57 / 0.701 & 34.28 / 0.951 & 34.68 / 0.950 & 38.68 / \underline{0.961} & \underline{38.89} / 0.957 & \textbf{39.03} / \textbf{0.965} \\
 & L096 & 25.93 / 0.715 & 35.41 / 0.938 & 35.78 / 0.941 & \underline{37.05} / 0.957 & 36.86 / \underline{0.959} & \textbf{37.24} / \textbf{0.967} \\
 & L109 & 26.73 / 0.677 & 33.84 / 0.932 & 35.34 / 0.964 & 40.29 / 0.973 & \underline{40.40} / \textbf{0.983} & \textbf{40.55} / \underline{0.976} \\
 & L143 & 25.79 / 0.726 & 35.65 / 0.966 & 36.26 / 0.958 & 37.57 / 0.969 & \underline{37.75} / \underline{0.973} & \textbf{37.92} / \textbf{0.979} \\
 & L192 & 26.65 / 0.704 & 36.98 / 0.956 & 37.32 / 0.967 & 38.98 / \underline{0.976} & \underline{39.23} / 0.974 & \textbf{39.37} / \textbf{0.981} \\
 & L506 & 26.38 / 0.653 & 36.82 / 0.969 & 37.13 / 0.963 & 39.23 / 0.971 & \underline{39.47} / \textbf{0.981} & \textbf{39.69} / \underline{0.980} \\
 & Avg. & 26.18 / 0.696 & 35.50 / 0.952 & 36.09 / 0.957 & 38.63 / 0.968 & \underline{38.77} / \underline{0.971} & \textbf{38.97} / \textbf{0.975} \\
\hline\hline
\end{tabular}
\vspace{-19pt}
\end{table*}

The proposed method was evaluated on simulated clinical datasets and real-world biological specimens. High-quality CT volumes from the AAPM Mayo Clinic Low-Dose CT Grand Challenge were utilized as ground truth and voxelized to a $256 \times 256 \times 256$ spatial resolution. To emulate CBCT acquisition, digital reconstructed radiographs were generated from these phantoms to produce $512 \times 800$ planar projections. Real-world biological projection data were additionally included to assess practical robustness.

RGS was implemented in PyTorch and optimized over 25,000 iterations. The base component was initialized with 50,000 large-scale Gaussian primitives, while the detail component utilized 30,000 smaller primitives (initial opacity 0.01) localized to regions possessing the top 5\% high-frequency spectral energy. Optimization followed a two-phase schedule: 5,000 warm-up iterations with a frozen detail component, followed by joint refinement utilizing a low-frequency consistency loss weight of 0.5 and a base learning rate decay factor of 0.1. The spatial position learning rate was exponentially annealed from $2 \times 10^{-3}$ to $2 \times 10^{-6}$, while learning rates for density, scaling, and rotation were fixed at $1 \times 10^{-3}$ alongside periodic adaptive density control.

\subsection{Comparative Evaluation}
\label{sec:comparative_eval}

To evaluate RGS, we benchmark it against analytical (FDK), iterative (SART), and recent neural rendering methods (NAF, R$^2$-Gaussian, and 3DGR). Evaluations utilize the clinical datasets, focusing on complex anatomical regions like the chest and abdomen under ultra sparse-view configurations. Reconstruction quality is quantitatively assessed via PSNR and SSIM.

\subsubsection{Quantitative Results}
\label{sec:quant_results}

Table \ref{tab:clinical_results} presents the quantitative comparison of all evaluated methods on the clinical AAPM dataset across varying sparse-view configurations (60, 40, and 20 views). The results demonstrate that the proposed RGS framework consistently yields the highest performance across all metrics and sparsity levels.

As expected, the analytical FDK algorithm exhibits severe performance degradation under ultra-sparse-view settings. While the iterative SART method provides better noise suppression than FDK, it tends to over-smooth structural boundaries, resulting in suboptimal SSIM scores. Among the neural rendering baselines, NAF, R$^2$-Gaussian, and 3DGR show progressive improvements in global geometric reconstruction. However, their performance plateaus when recovering complex textures, reflecting the inherent spectral bias of standard photometric optimization.

In contrast, by explicitly compensating for high-frequency residuals via the detail component, RGS achieves better voxel-wise accuracy and structural fidelity. Notably, under the highly ill-posed 20-view setting, our method maintains a PSNR of 34.68 dB and an SSIM of 0.909, outperforming the second-best baseline. This consistent quantitative gain across decreasing view counts validates the robustness and effectiveness of the proposed spectral-spatial collaborative optimization strategy.

\subsubsection{Visual Comparison}

To qualitatively evaluate reconstruction fidelity under ultra view sparsity, we present a visual comparison of the chest and abdomen regions reconstructed from 20 views in \cref{fig:visual_comparison_20views}. The chest slice highlights fine lung markings and vascular structures, while the abdomen slice features low-contrast soft tissues and complex trabecular bone patterns.

As illustrated in \cref{fig:visual_comparison_20views}(a), the analytical FDK method is severely degraded by streak artifacts radiating from high-density structures, which completely obscure critical anatomical boundaries. While the iterative SART and neural rendering baselines successfully suppress these artifacts and restore global topological structures, they consistently exhibit noticeable texture over-smoothing. This degradation is a direct consequence of the spectral bias inherent in standard photometric optimization; these models prioritize low-frequency structural approximations at the expense of high-frequency variance. Consequently, fine vascular terminals in the lung parenchyma are blurred, and the porous trabecular texture within the vertebra is falsely rendered as a homogeneous block.

In contrast, the proposed RGS framework demonstrates superior detail preservation without reintroducing noise or artifacts. By explicitly compensating for high-frequency projection residuals through the detail component, our method effectively resolves the fine lung vascularity and the spongy texture of the trabecular bone. The preserved structural sharpness and textural fidelity confirm that the spectral-spatial collaborative optimization successfully mitigates the inherent trade-off between artifact reduction and detail recovery, yielding visual results most consistent with the ground truth.

Ablation studies are conducted to evaluate the individual contributions of the core RGS components. Specifically, the impacts of the wavelet-based spectral prior and the curriculum-driven collaborative optimization strategy are assessed via quantitative metrics and qualitative frequency analyses.

\begin{figure}[htbp]
    \centering
    \includegraphics[width=\columnwidth]{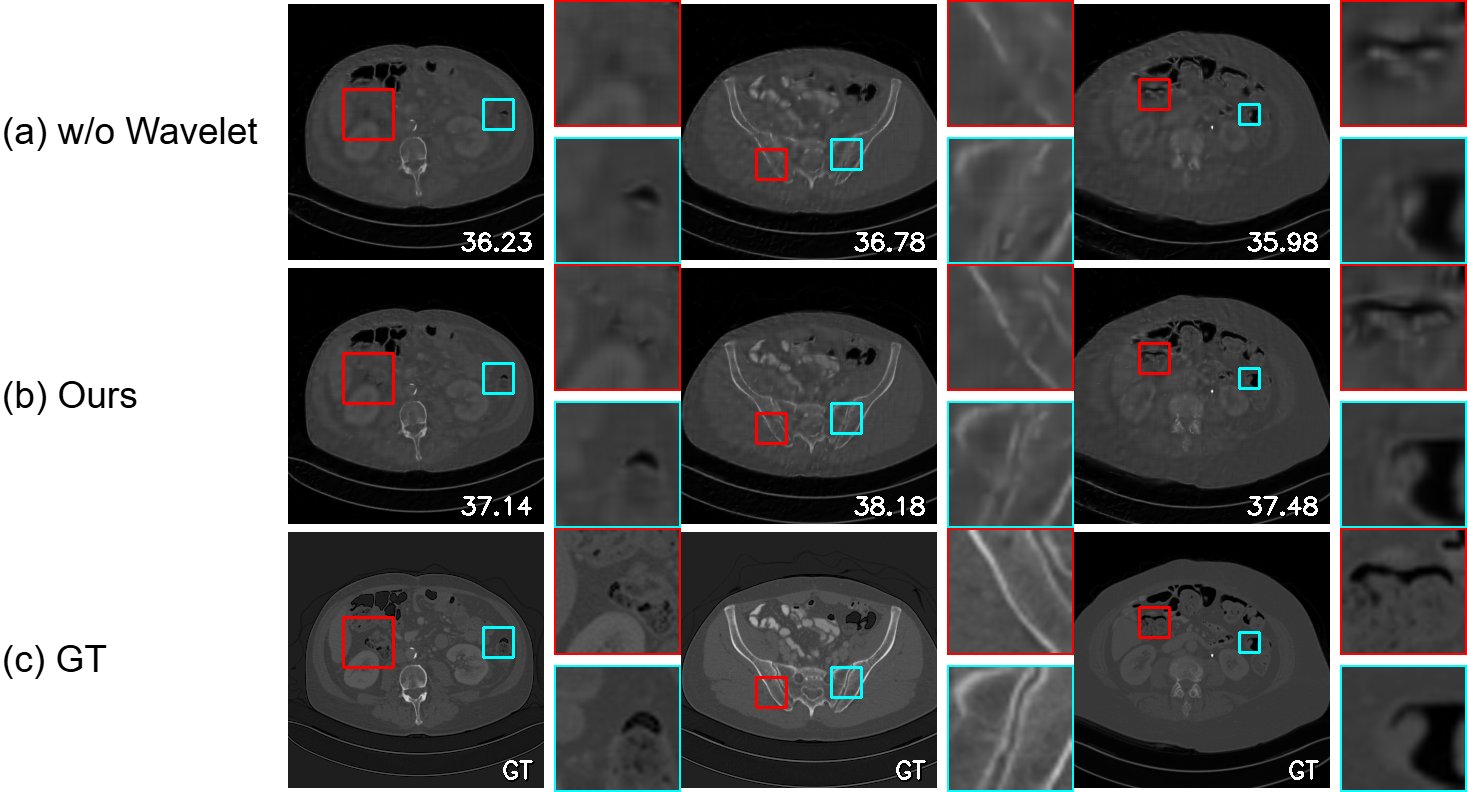} 
    \vspace{-4mm}
    \caption{Visual ablation study on the wavelet-based spectral prior under a 40-view sparse configuration. (a) Reconstruction by the model variant without spectral prior guidance (w/o Wavelet). (b) Reconstruction by the complete framework (RGS). (c) GT.}
    \label{fig:ablation_w/o_wavelet_40views}
    \vspace{-20pt}
\end{figure}

\subsection{Ablation Studies}
\label{sec:ablation}
\subsubsection{Effectiveness of Spectral Prior Guidance}

\begin{figure}[htbp]
    \centering
    \includegraphics[width=\columnwidth]{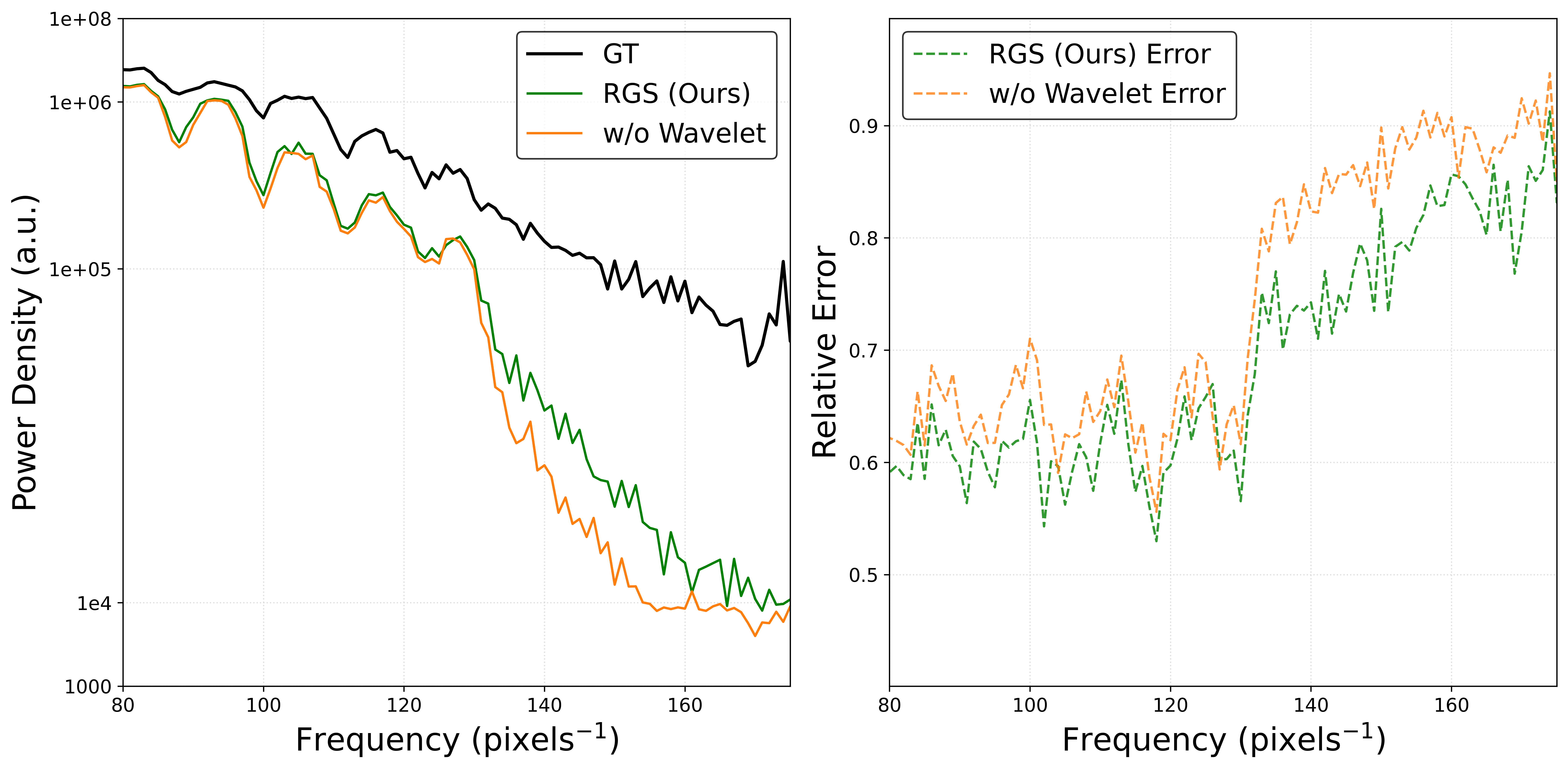} 
    \vspace{-4mm}
    \caption{Quantitative frequency analysis evaluating the effectiveness of the wavelet module, displaying the radially averaged power density (left) and the corresponding relative error with respect to the ground truth (right).}
    \label{fig:ablation_frequency}
\end{figure}

Removing the spectral prior guidance (w/o Wavelet) yields a consistent drop in both PSNR and SSIM across all volumes, as reported in \cref{tab:ablation_quant}, and manifests visually as over-smoothed and blurred structural boundaries in \cref{fig:ablation_w/o_wavelet_40views}. The frequency analysis in \cref{fig:ablation_frequency} confirms that without explicit high-frequency priors, the model succumbs to spectral bias. While low-frequency performance remains comparable, the ablated variant exhibits a precipitous energy decay and a corresponding surge in relative error at higher spatial frequencies. Conversely, the full RGS model tracks the ground truth power density across the entire spectrum, confirming that wavelet-derived priors are indispensable for preserving micro-structural fidelity.

\begin{table}[htbp]
    \centering
    \vspace{-5pt}
    \caption{Quantitative ablation study of the spectral prior on clinical datasets under a 40-view sparse configuration.}
    \label{tab:ablation_quant}
    \setlength{\tabcolsep}{10pt}
    \begin{tabular}{lcccc}
        \toprule
        \multirow{2}{*}{\textbf{Dataset}} & \multicolumn{2}{c}{\textbf{w/o Wavelet}} & \multicolumn{2}{c}{\textbf{RGS (Ours)}} \\
        \cmidrule(lr){2-3} \cmidrule(lr){4-5}
         & PSNR $\uparrow$ & SSIM $\uparrow$ & PSNR $\uparrow$ & SSIM $\uparrow$ \\
        \midrule
        L067 & 36.18 & 0.942 & 37.76 & 0.960 \\
        L096 & 35.15 & 0.916 & 36.01 & 0.932 \\
        L109 & 36.25 & 0.951 & 37.88 & 0.962 \\
        \midrule
        \textbf{Average} & 35.86 & 0.936 & 37.22 & 0.951 \\
        \bottomrule
    \end{tabular}
    \vspace{-5pt}
\end{table}

\subsubsection{Necessity of Curriculum Optimization}

\begin{figure}[htbp]
    \vspace{-10pt}
    \centering
    \includegraphics[width=\columnwidth]{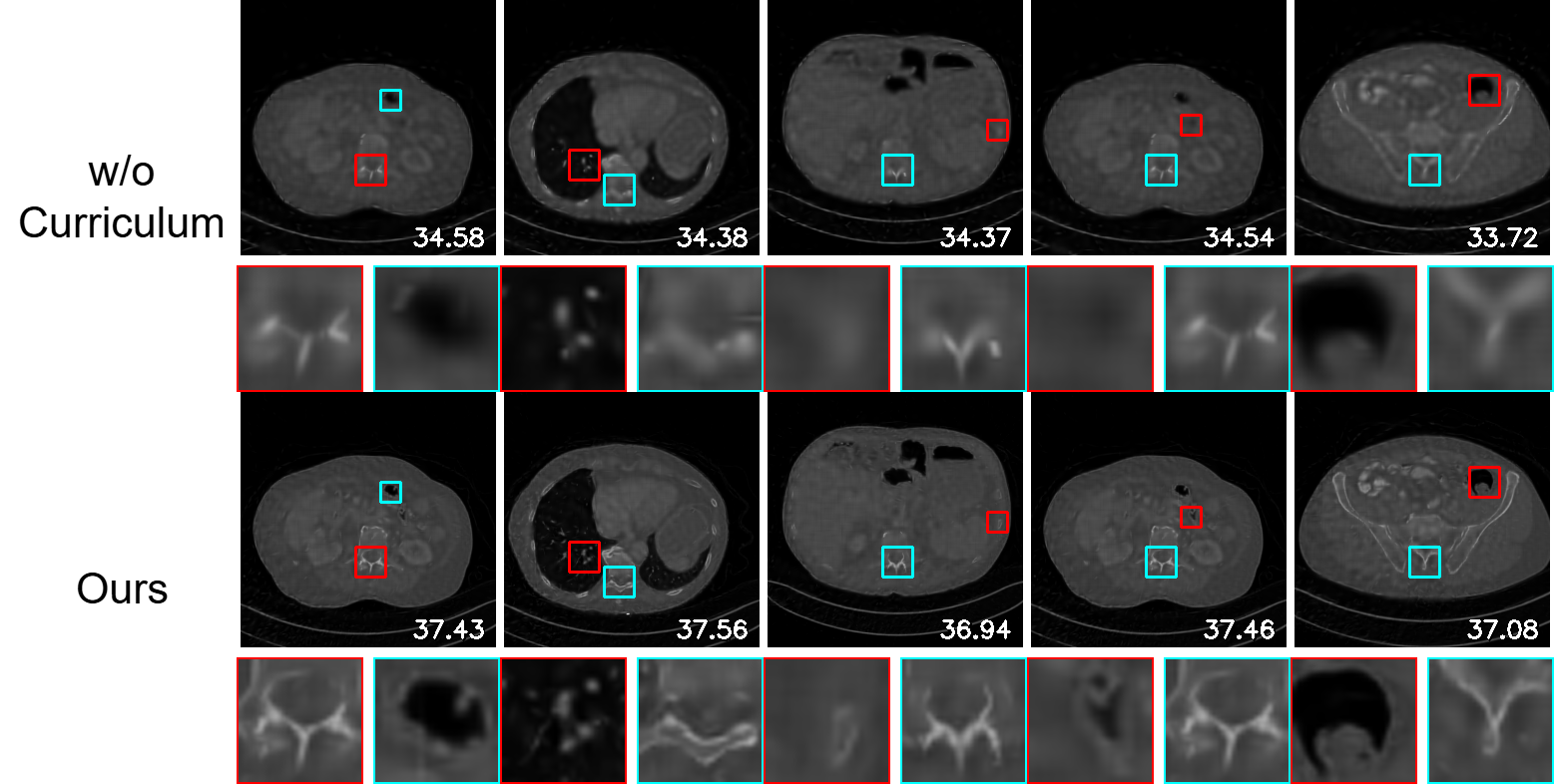} 
    \vspace{-4mm}
    \caption{Visual ablation study evaluating the impact of curriculum optimization under a 40-view sparse configuration, comparing the model without temporal decoupling (w/o Curriculum) against the complete framework (Ours).}
    \label{fig:ablation_curriculum}
    \vspace{-10pt}
\end{figure}

Bypassing the low-frequency warm-up phase to train both Gaussian sets simultaneously (w/o Curriculum) induces a substantial degradation in voxel-wise accuracy and structural similarity, as reported in \cref{tab:ablation_curriculum}. Visually, this simultaneous optimization causes a systematic failure to resolve fine structural boundaries in \cref{fig:ablation_curriculum}. The fundamental mechanism driving this failure is unconstrained spectral competition. Without establishing a stable low-frequency topological anchor first, the base and detail components aggressively compete to explain the same projection residuals. The base component, possessing larger spatial scale and dominant initial energy, excessively parameterizes the volumetric space during early iterations. It absorbs the high-frequency variance allocated to the detail component, effectively suppressing texture recovery. This spectral crosstalk disrupts the intended functional decoupling, underscoring that the curriculum-driven schedule is essential to mitigate component interference and ensure stable convergence.

\begin{table}[htbp]
    \vspace{-5pt}
    \centering
    \caption{Quantitative ablation study of the curriculum optimization strategy on clinical datasets under a 40-view sparse configuration.}
    \label{tab:ablation_curriculum}
    \setlength{\tabcolsep}{10pt}
    \begin{tabular}{lcccc}
        \toprule
        \multirow{2}{*}{\textbf{Dataset}} & \multicolumn{2}{c}{\textbf{w/o Curriculum}} & \multicolumn{2}{c}{\textbf{Ours}} \\
        \cmidrule(lr){2-3} \cmidrule(lr){4-5}
         & PSNR $\uparrow$ & SSIM $\uparrow$ & PSNR $\uparrow$ & SSIM $\uparrow$ \\
        \midrule
        L143 & 34.15 & 0.912 & 36.92 & 0.941 \\
        L192 & 34.60 & 0.920 & 37.55 & 0.950 \\
        L506 & 35.20 & 0.935 & 38.41 & 0.968 \\
        \midrule
        \textbf{Average} & 34.65 & 0.922 & 37.62 & 0.953 \\
        \bottomrule
    \end{tabular}
\end{table}

In summary, the ablation experiments systematically validate the individual and synergistic contributions of the proposed architectural modules.

\subsection{Evaluation on Real-World Biological Specimen}

To further validate the practical applicability and generalization capability of the proposed RGS framework beyond simulated clinical datasets, we conducted an evaluation using real-world tomographic data. A physical rhinoceros beetle specimen was scanned using a CT system, and 40 planar projections were uniformly acquired over a 360$^\circ$ rotational trajectory to construct a challenging sparse-view reconstruction scenario. 

\begin{figure}[htbp]
    \centering
    \includegraphics[width=\columnwidth]{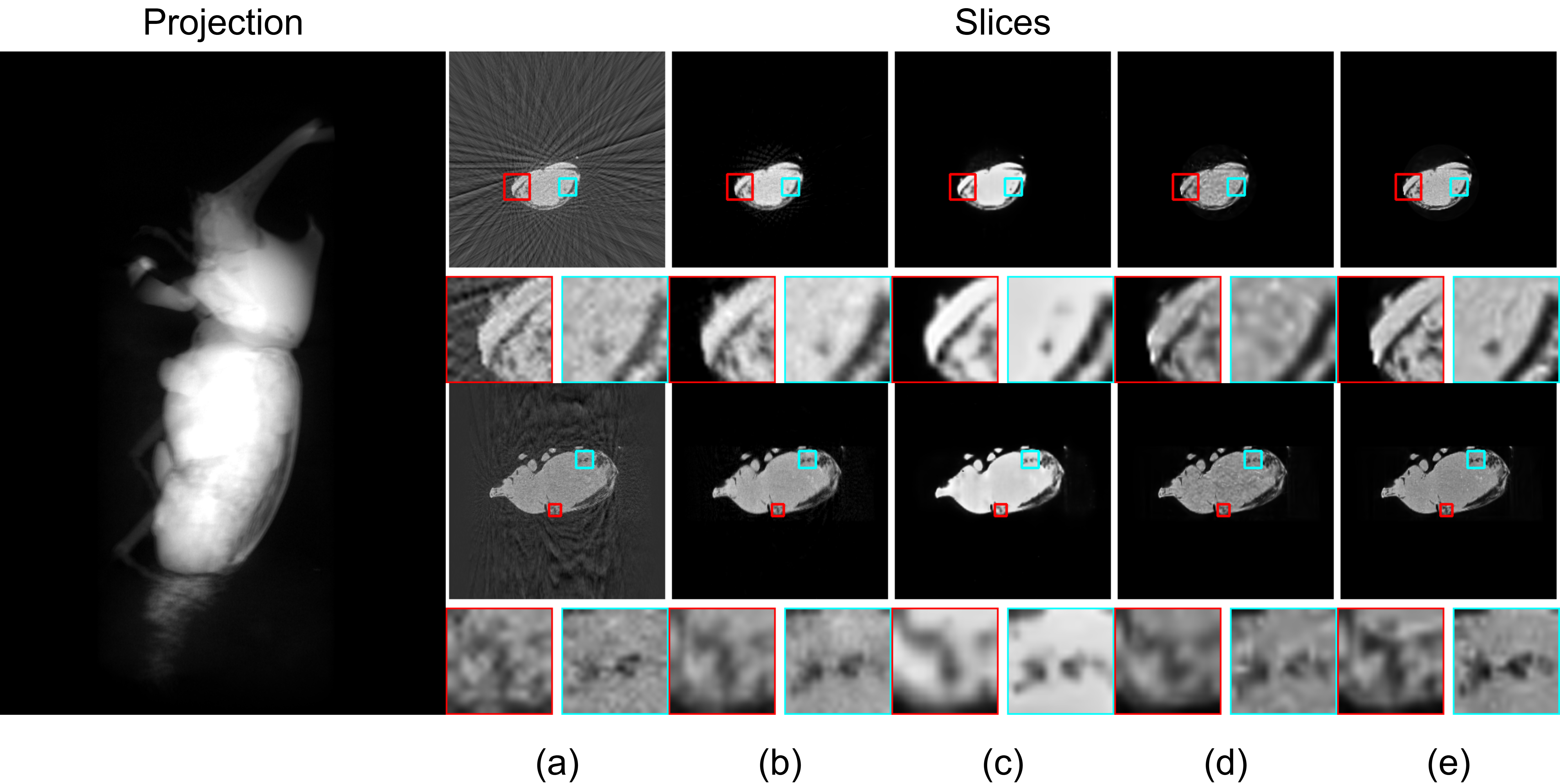}
    \vspace{-18pt}
    \caption{Visual comparison of reconstructed slices for a real rhinoceros beetle specimen under a 40-view sparse configuration. The leftmost panel displays a sample raw projection. (a) FDK, (b) SART, (c) NAF, (d) R$^2$-Gaussian, and (e) Ours.}
    \label{fig:beetle_evaluation}
    \vspace{-10pt}
\end{figure}

\cref{fig:beetle_evaluation} presents a visual comparison of the reconstructed cross-sectional slices alongside magnified regions of interest. The leftmost panel displays a raw acquired projection, highlighting the complex internal morphology and varying density of the biological specimen. As observed in the reconstruction results, the analytical FDK result in \cref{fig:beetle_evaluation}(a) is severely degraded by undersampling artifacts and pervasive streak noise, which obscure fine structural boundaries. The iterative SART result in \cref{fig:beetle_evaluation}(b) effectively suppresses the background noise but introduces noticeable over-smoothing, particularly blurring the sharp edges of the chitinous exoskeleton. While neural rendering baselines such as the NAF result in \cref{fig:beetle_evaluation}(c) and the R$^2$-Gaussian result in \cref{fig:beetle_evaluation}(d) manage to restore the global topological structure, they remain constrained by spectral bias, failing to adequately resolve the high-frequency textural details of the internal micro-structures.
\vspace{-10pt}

In contrast, our proposed approach, shown in \cref{fig:beetle_evaluation}(e), explicitly integrates spectral prior guidance and residual compensation, successfully overcoming these over-smoothing limitations. As highlighted by the red and cyan bounding boxes, RGS accurately recovers the delicate internal organ textures and preserves the sharp, high-contrast boundaries of the outer shell. This experiment empirically confirms that the proposed method maintains robust detail preservation and geometric fidelity when applied to raw physical acquisition data, demonstrating significant potential for high-precision biological and pre-clinical imaging applications.

\section{Conclusion}

This work presented RGS to overcome the spectral bias inherent in explicit neural rendering for ultra sparse-view CBCT reconstruction. By resolving the fundamental incompatibility between non-negative physical attenuation and signed wavelet coefficients, we introduced a spectrally-decoupled gaussian representation that stratified the volumetric field into a geometric base and a residual detail component. Guided by a curriculum-based collaborative optimization strategy, this framework effectively translated explicit high-frequency spectral priors into physically consistent implicit residual compensation. Consequently, RGS successfully reconciled the trade-off between global artifact suppression and local textural preservation, demonstrating superior structural fidelity over existing baselines.

\bibliographystyle{IEEEtran} 
\bibliography{refs}          

\end{document}